\pdfoutput=1

\documentclass[11pt]{article}

\usepackage[]{acl}

\usepackage{times}
\usepackage{latexsym}

\usepackage[T1]{fontenc}

\usepackage[utf8]{inputenc}

\usepackage{microtype}
\usepackage{booktabs}
\usepackage{graphicx}
\usepackage{amssymb}

\usepackage{array}
\newcolumntype{P}[1]{>{\centering\arraybackslash}p{#1}}
\newcolumntype{L}[1]{>{\raggedleft\arraybackslash}p{#1}}
\newcolumntype{R}[1]{>{\raggedright\arraybackslash}p{#1}}

\title{Chain of Logic: Rule-Based Reasoning with Large Language Models}

\author{
    \begin{tabular}{c}
    Sergio Servantez$^\dagger$ \quad Joe Barrow$^\blacktriangledown$$^\ddag$ \quad Kristian Hammond$^\dagger$ \quad Rajiv Jain$^\ddag$ \vspace{.5mm} \\
    \end{tabular}
    \\ \vspace{.5mm}
    \small
    \begin{tabular}{c}
    $^\dagger$Northwestern University \quad$^\ddag$ Adobe Research \quad$^\blacktriangledown$ Pattern Data \\
    \end{tabular}
    \\ \vspace{.5mm}
    \small
    \begin{tabular}{c}
    \texttt{servantez@u.northwestern.edu} \quad \texttt{joe.barrow@patterndataworks.com} \quad \\ \texttt{Kristian.Hammond@northwestern.edu} \quad \texttt{rajijain@adobe.com}
    \end{tabular}
    \vspace{2mm} \\
}

\begin{document}
\maketitle
\begin{abstract}
Rule-based reasoning, a fundamental type of legal reasoning, enables us to draw conclusions by accurately applying a rule to a set of facts. We explore causal language models as rule-based reasoners, specifically with respect to compositional rules - rules consisting of multiple elements which form a complex logical expression. Reasoning about compositional rules is challenging because it requires multiple reasoning steps, and attending to the logical relationships between elements. We introduce a new prompting method, Chain of Logic, which elicits rule-based reasoning through decomposition (solving elements as independent threads of logic), and recomposition (recombining these sub-answers to resolve the underlying logical expression). This method was inspired by the IRAC (Issue, Rule, Application, Conclusion) framework, a sequential reasoning approach used by lawyers. We evaluate chain of logic across eight rule-based reasoning tasks involving three distinct compositional rules from the LegalBench benchmark and demonstrate it consistently outperforms other prompting methods, including chain of thought and self-ask, using open-source and commercial language models.
\end{abstract}

\section{Introduction}
\label{sec:intro}

\begin{figure}
  \centering
  \includegraphics[width=\linewidth]{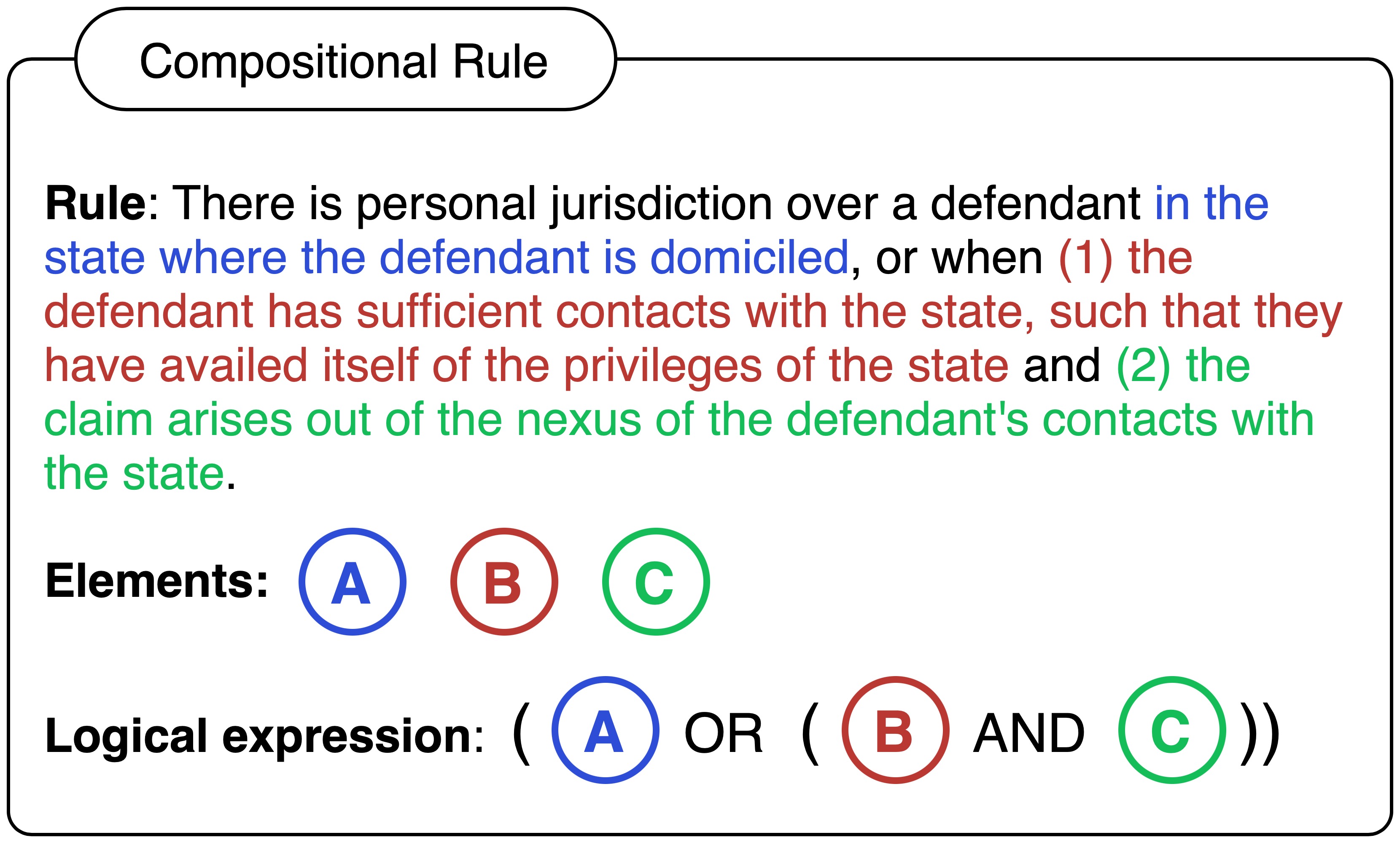}
  \caption{Example showing compositional structure of rule for Personal Jurisdiction task from LegalBench. Color coding is used to identify rule elements and illustrate how these elements form a complex logical expression. Reasoning about compositional rules requires not only correctly applying each element to a fact pattern, but also resolving the logical expression. If the logical expression evaluates to true, it triggers a consequence (personal jurisdiction exists).}
  \label{figure:intro} 
\end{figure}

The surge in reasoning capabilities of language models (LMs) has the potential to transform the legal industry, a sector inherently reliant on sophisticated reasoning. The development of models that can assist and validate the reasoning processes of legal practitioners promises to usher in a new legal era. Such advancements would not only enhance the efficiency of legal services but also expand the capacity of professionals to service more clients, thereby broadening access to justice. At present, language models are prone to hallucinations in a legal setting \cite{dahl2024large}, and even powerful LMs like GPT-4 struggle to perform basic legal tasks \cite{blairstanek2023blt}.

Legal tasks typically require sophisticated rule-based reasoning. These rules are written in natural language and expressed in many forms, including statutes, judicial holdings and even contract provisions. Similar to an if/then statement, a rule has an antecedent (a condition that can be evaluated to true or false) and a consequent (the outcome triggered if the antecedent is satisfied). Even the earliest recorded laws, written in Sumerian on clay tablets, have used this conditional structure \cite{roth1995law}. Rule-based reasoning allows us to draw conclusions by applying a rule to a set of facts to determine if these preconditions are satisfied. For example, if parking is prohibited (consequent) between 2pm and 4pm (antecedent), and we know it is currently 3pm, then we can conclude that parking is currently prohibited. Often rules, especially complex rules, are compositional in nature meaning the antecedent consists of multiple conditions joined by \textit{and} and \textit{or} operators forming a complex logical expression (see Figure \ref{figure:intro}). In law, these constituent conditions are called rule elements. Recent work demonstrates that models can struggle to reason about even basic compositional rules \cite{10.1145/3594536.3595163}.

\begin{figure*}[t]
  \centering
  \includegraphics[width=\linewidth]{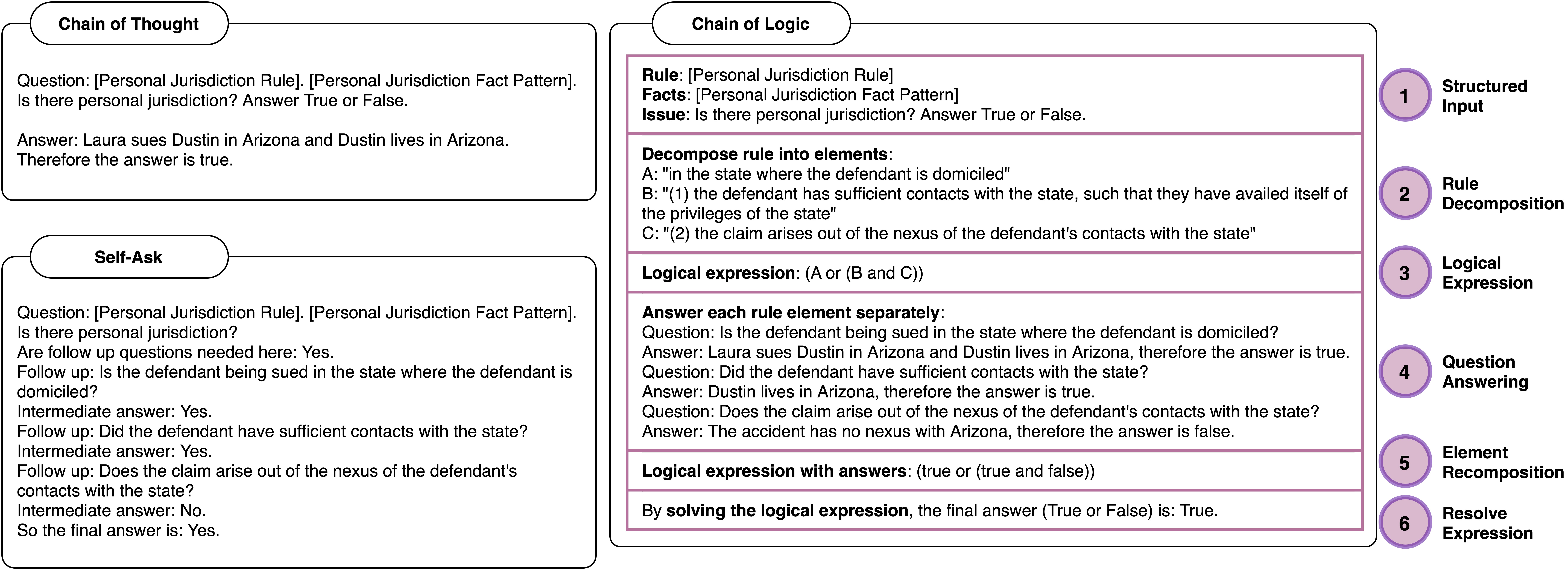}
  \caption{Comparing one-shot examples demonstrating chain of thought, self-ask and our chain of logic method on the Personal Jurisdiction task. Through a sequence of reasoning steps, chain of logic decomposes a rule into elements which are solved independently, before recomposing sub-answers to arrive at a final conclusion. See Section \ref{chain_of_logic} for a detailed discussion on the chain of logic approach.}
  \label{figure:methods}
\end{figure*}

We explore large language models as rule-based reasoners, evaluating several common in-context learning methods across 8 tasks and 3 distinct rules from LegalBench, a collaboratively constructed legal reasoning benchmark \cite{guha2023legalbench}. LegalBench tasks were designed by subject matter experts to evaluate useful legal reasoning skills. Prior work on LegalBench has focused on evaluating the capacity of models to perform rule-based reasoning given in-context demonstrations of the same rule being applied to varying fact patterns. These few-shot prompts are valuable in law where annotated data is scarce, limiting the ability to fine-tune models for specific legal tasks. Yet requiring several reasoning examples for each rule in a real world setting poses a challenge in terms of both cost and scalability. Additionally, a same-rule setting allows the model to largely mimic the reasoning path of the in-context examples. In contrast, a different-rule setting enables us to examine a model's ability to dynamically generate this reasoning path - the same way a lawyer would reason. Our work shifts focus to evaluating and improving a model's ability to perform rule-based reasoning given only a \emph{single} in-context demonstration of a \emph{different} rule application, thus removing the need to store examples for each rule. 

In this work, we introduce \textbf{Chain of Logic}, a new prompting approach to guide LMs in reasoning about compositional rules by decomposing rule elements into a series of logical statements (evaluate to true or false), before explicitly recomposing these element answers to resolve the underlying logical expression. Understanding natural language questions requires understanding how to break down a question into a set of steps yielding an answer \cite{wolfson2020break}. The inspiration for our method is rooted in the IRAC Framework, an approach to legal reasoning where lawyers break down the reasoning process into four sequential steps: issue spotting, rule identification, rule application and conclusion. During the rule application step, rules are typically divided into their elements and addressed separately before recombining these threads to arrive at a conclusion. In devising our chain of logic approach, our interdisciplinary team combined this domain knowledge with a variety of insights from prior work on in-context learning.

Our experiments show chain of logic outperforms other prompting methods, including chain of thought \cite{wei2023chainofthought} and self-ask \cite{press2023measuring}, across a variety of rule-based reasoning tasks using both open-source and commercial models. We demonstrate that given a single example of chain of logic, a model can learn to generalize this approach to a different rule and fact pattern, and thereby improve its reasoning capabilities. 

\begin{figure*}[t]
  \centering
  \includegraphics[width=\linewidth]{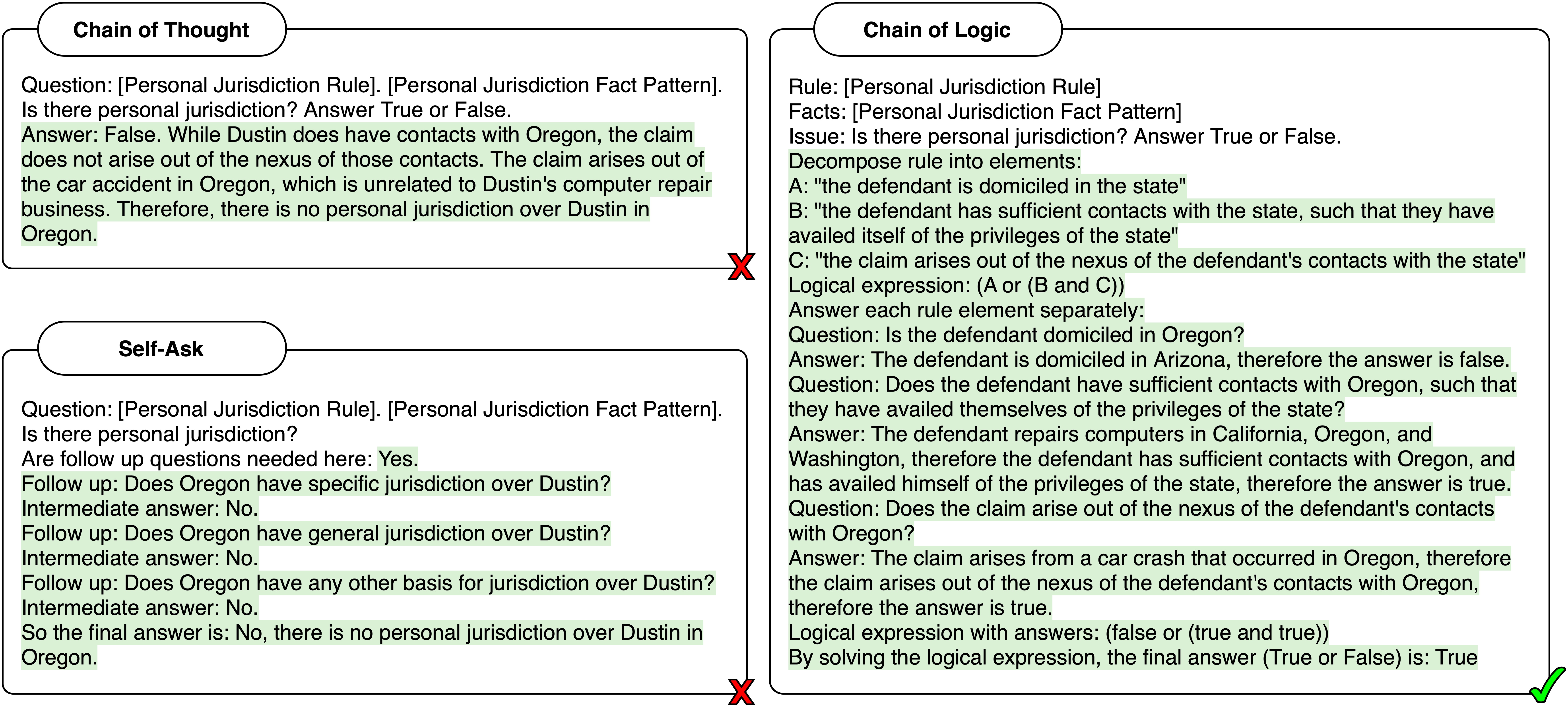}
  \caption{Comparing GPT-3.5 output for chain of thought, self-ask and our chain of logic method on the same Personal Jurisdiction task. Chain of logic prompting elicits large language models to reason about complex rules while also constructing an interpretable reasoning path. The prompts here are abridged, omitting a one-shot example for each method from the Diversity Jurisdiction task (see Section \ref{tasks}).}
  \label{figure:output}
\end{figure*}

\section{Background}

Large language models have demonstrated strong capabilities as zero-shot \cite{kojima2023large} and few-shot \cite{brown2020language} reasoners. Chain of thought prompting enhanced problem solving performance even further by eliciting models to produce intermediate reasoning steps \cite{wei2023chainofthought}. This approach proved effective at solving complex tasks, including arithmetic and commonsense reasoning tasks. Many other prompting methods followed with a similar aim of decomposing a complex task into a sequence of simpler subtasks. Self-ask improved multi-hop question answering performance beyond chain of thought prompting by guiding the model to explicitly pose and answer intermediate questions \cite{press2023measuring}. Similar to decomposing a multi-hop question into a sequence of intermediate questions, compositional rules can be decomposed into rule elements. However, reasoning over these rules requires not only answering each element correctly, but also understanding the logical relationships that exists between these elements. For example, if a rule with two elements resolves to (true, false), then the final answer depends on whether the boolean relationship between those elements is \emph{and} or \emph{or}. In Section x, we show existing prompting methods can resolve rule elements correctly while still getting the final answer incorrect. This motivates the need for a more robust prompting method which attends to both the element level answers and the underlying logical structure. Further motivating this work, zero-shot methods have been observed outperforming one-shot and few-shot approaches on legal reasoning tasks \cite{yu2022legal, 10.1145/3594536.3595163}, suggesting models can struggle with in-context learning in a legal setting.

\section{Chain of Logic} \label{chain_of_logic}

We introduce chain of logic, a new prompting approach to elicit rule-based reasoning in LMs through a series of instructive reasoning steps. Each step in this series helps inform the next, and enables the model to unravel the many reasoning tasks needed to arrive at the right conclusion. Our method builds on chain of thought and self-ask by not only considering problem decomposition, but also attending to relationships between subtasks that can dictate the final answer, which we later show improves performance. Our prompt begins with a single example demonstrating the chain of logic approach, appended with the inference-time problem. Figure \ref{figure:methods} contains a one-shot example of our approach depicting the sequence of reasoning steps which we describe below:

\begin{itemize}
  \item \textbf{Step 1:} \underline{Structured Input}. Clearly label the task inputs to identify rule, facts and issue, similar to the IRAC Framework.
  \item \textbf{Step 2:} \underline{Rule Decomposition}. Decompose the rule into elements by identifying relevant text spans and assigning elements to variables. Similar to the least-to-most approach \cite{zhou2023leasttomost}, our method first identifies the subtasks before trying to solve them.
  \item \textbf{Step 3:} \underline{Logical Expression}. Construct a complex logical expression representing the logical relationships that exist between these element variables.
  \item \textbf{Step 4:} \underline{Question Answering}. Iterate through each rule element, rephrasing the element as a question before providing a rationale and answer. Like self-ask, we dynamically construct this series of sub-questions, but for our approach this construction is informed by the elements identified in step 2. The model need only rephrase each element as question.
  \item \textbf{Step 5:} \underline{Element Recomposition}. Replace element variables in the logical expression with the corresponding sub-answers from the previous step.
  \item \textbf{Step 6:} \underline{Resolve Expression}. Resolve the logical expression populated with sub-answers to arrive at the final solution.
\end{itemize}

During inference, the model performs each step in this process automatically given only a one-shot example and the inference task inputs (formatted as described in step 1). Since the one-shot task inputs (rule, facts, issue) are different from the task inputs being tested, the model is reasoning on its own at each step, including deciding how many rule elements exist, the text span of each element and the logical relationships between them. This more closely resembles a real world setting where lawyers apply the IRAC framework to a rule and fact pattern being observed for the first time.

The intermediate steps in our approach provide a detailed accounting of each factor contributing to the final conclusion (see Figure \ref{figure:output} for an example of model output). This allows for transparent and explainable decision-making, which is particularly vital in the legal domain where the justification for an answer can be as important as the answer itself. Chain of logic also enables us to debug inaccurate conclusions by retracing the reasoning path to locate errors in rule application and logic.

\section{Experiments}

We conduct a series of experiments to compare our proposed chain of logic approach with existing prompting methods, and to evaluate the rule-based reasoning capabilities of LMs in a different-rule setting. While we sometimes observe an inverse relationship between prompt complexity and performance (simpler prompts perform better), we find that chain of logic consistently outperforms other prompting methods across a range of rule-based reasoning tasks. 

\begin{table*}[t]
  \centering
  \begin{tabular}{ R{62mm} L{14mm} L{14mm} L{14mm} L{14mm} L{14mm} }
    \toprule
    & GPT-3.5 & GPT-4 & Llama-2 & Mistral & Average \\ \midrule
    Zero-Shot & 76.3 & 90.4 & 74.2 & 60.8 & 75.4\\      
    Zero-Shot-LR \cite{yu2022legal} & 57.2 & 90.6 & 65.3 & 62.8 & 69.0 \\ 
    Zero-Shot-LS \cite{jiang2023legal} & 75.5 & 90.1 & 58.4 & 52.2 & 69.1\\ \midrule

    Standard Prompting \cite{brown2020language} & 72.6 & 88.6 & 65.1 & 45.5 & 68.0\\
    Chain of Thought \cite{wei2023chainofthought} & 70.1 & 89.1 & 71.8 & 46.4 & 69.4\\
    Self-Ask \cite{press2023measuring} & 68.2 & 86.3 & 67.5 & 46.5 & 67.1\\
    Chain of Logic (ours) & \textbf{87.0} & \textbf{92.3} & \textbf{74.6} & \textbf{63.1} & \textbf{79.3}\\
    \bottomrule
  \end{tabular}
  \caption{Aggregated performance (accuracy) across all rules, including Personal Jurisdiction, Diversity Jurisdiction and J. Crew Blocker, using both open-source and commercial language models. Chain of logic consistently outperforms other prompting methods. All methods are one-shot, except zero-shot approaches.}
  \label{tab:results}
\end{table*}

\subsection{Tasks} \label{tasks}

LegalBench \cite{guha2023legalbench} is a legal reasoning benchmark designed and constructed by an interdisciplinary team of computer scientists and legal professionals. The tasks in this benchmark have been designed by subject matter experts to measure legal reasoning capabilities that are both interesting and useful. We consider 8 rule-based reasoning tasks involving 3 distinct compositional rules from this benchmark:

\begin{itemize}
\item {\textbf{Personal Jurisdiction}} (1 task): This task involves determining whether a court has the authority to preside over a dispute based on where the defendant is domiciled, and whether the defendant had sufficient contacts with the forum state and the claim arose out of the nexus of those contacts. There are 50 test samples for this task.
\item {\textbf{Diversity Jurisdiction} (6 tasks)}: This task involves determining whether a federal court can preside over a lawsuit pertaining to state law based on whether complete diversity exists (no plaintiff and defendant are citizens of the same state) and the amount in controversy exceeding \$75,000. There are 6 diversity jurisdiction tasks with increasingly complex fact patterns, from Diversity Jurisdiction 1 (easiest) to Diversity Jurisdiction 6 (hardest). There are 300 test samples for each diversity jurisdiction task.
\item {\textbf{J.Crew Blocker}} (1 task): This task involves determining whether a provision from a loan agreement contains a J.Crew Blocker restrictive covenant based on whether the provision prohibits transferring IP to an unrestricted subsidiary or requires lender consent for IP transfers to a subsidiary. There are 54 test samples for this task.
\end{itemize}

Each sample from all tasks contains a rule, fact pattern and question. To correctly answer the question, the rule must be applied to the fact pattern. The prompt for each task contains a one-shot example from another rule, followed by the test sample. The benchmark expects an answer of yes or no for each question. For our method we reformat the answer to true or false to more closely resemble formal logic. If the model's response is ambiguous, we use a second prompt "Therefore the answer (true or false) is", following Kojima et al. \cite{kojima2023large}.

\subsection{Baseline Methods}

In addition to our chain of logic approach (see Section \ref{chain_of_logic}), we examine six prompting methods. We first explore  zero-shot prompting where the prompt includes only the test sample (rules, facts, question) with no in-context demonstrations. Blair-Stanek et al. showed that zero-shot prompting can outperform few-shot methods for some legal reasoning tasks, even with chain of thought reasoning \cite{10.1145/3594536.3595163}. This suggests some models could have difficulty learning through in-context demonstrations in a legal setting. We include zero-shot prompting in our experiments to further explore this hypothesis. For completeness, we also include two zero-shot methods designed for legal reasoning tasks from prior work. Legal syllogism (LS) prompting is a zero-shot approach to legal judgement prediction where the prompt first defines legal syllogism before instructing the model to perform syllogistic reasoning \cite{jiang2023legal}. Legal reasoning (LR) prompting \cite{yu2022legal} is also a zero-shot method where the prompt includes a simple approach description: "Approach: Issue, rule, application, conclusion".

For the remaining methods, we use a one-shot approach where a single demonstration with correct answer is included in the prompt. First, we use standard prompting \cite{brown2020language} where the in-context demonstration includes only the sample and answer. Second, we explore chain of thought \cite{wei2023chainofthought} prompting which includes a rational written by a legal professional explaining the relevant reasoning before the demonstration answer. Last, we include the self-ask approach \cite{press2023measuring} where the one-shot example demonstrates explicitly asking and answering an intermediate question for each rule element before the final answer. See Figure \ref{figure:methods} for an illustration of these one-shot examples. While self-ask was originally evaluated on another compositional reasoning task, multi-hop question answering, its demonstrated ability to increase reasoning capacity by decomposing complex problems makes it a compelling method to explore in this work. 

\subsection{Language Models}

 We experiment with two commercial models, GPT-3.5-Turbo and GPT-4 \cite{openai2023gpt}, and two leading open-source models, Llama-2-70b-chat \cite{touvron2023llama} and Mistral-7B-OpenOrca, which is a Mistral-7b \cite{jiang2023mistral} model fine-tuned with Orca \cite{mukherjee2023orca}. We select open-source models of disparate size to investigate the relationship between model size and rule-based reasoning abilities. For all language models we set the temperature to 0.0 to make the output more focused and reproducible, and otherwise use default settings.

\begin{figure*}[t]
  \centering
  \includegraphics[width=\linewidth]{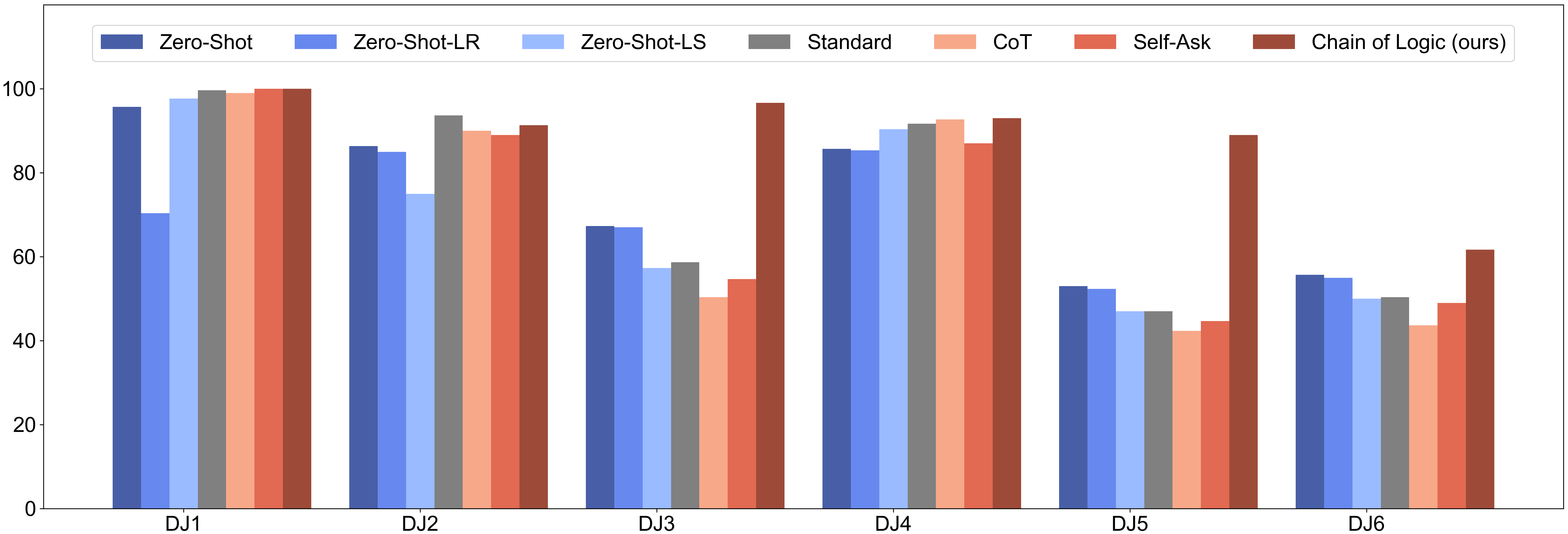}
  \caption{Accuracy (\%) across all 6 Diversity Jurisdiction tasks using GPT-3.5. The fact patterns in these tasks are increasingly complex, from DJ1 (easiest) to DJ6 (hardest). Chain of logic particularly outperforms other prompting methods for tasks requiring arithmetic operations (DJ3, DJ5, DJ6). See Section \ref{diversity_series} for a detailed discussion.}
  \label{figure:diversity}
\end{figure*}

\section{Results and Discussion}

Table \ref{tab:results} presents results for baseline methods and our chain of logic approach macro-averaged across all rule sets (Personal Jurisdiction, Diversity Jurisdiction and J.Crew Blocker), using commercial and open-source models. We report the macro-average here to provide a balanced view across rule sets, though as shown later in Figure \ref{diversity_series} and the appendix, Chain of logic performs best on the Diversity Jurisdiction task, which is downweighted with this metric. See the Appendix for more detailed model performance by rule. For each model, a zero-shot method is the best performing baseline. This is consistent with prior work finding that zero-shot methods can outperform one-shot and few-shot prompting on legal reasoning tasks \cite{yu2022legal, 10.1145/3594536.3595163}. Chain of logic significantly improves rule-based reasoning through in-context learning.

\textbf{For each model, chain of logic outperforms all baseline methods}, including zero-shot methods. The performance gap between chain of logic and the best performing baseline is 3.9\% on average. This gap is smaller for open source models (0.3\% and 0.4\%) and larger for commercial models, with GPT-3.5 having the largest performance improvement of 10.7\%. For this work we observe a large performance difference between the commercial and open source models, and hypothesize the performance gains may be larger for the stronger commercial models which can more easily follow the longer prompts. This can also be seen in the performance differences between the baseline zero and one-shot methods. For GPT-4 we believe the absolute performance gain may be lower due to ceiling effects of the model. The specialized zero-shot legal baselines demonstrate comparable performance to zero-shot for some models, but are also inconsistent. LR prompting slightly outperforms zero-shot using Mistral-7b and GPT-4, but also underperforms by 19.1\% for GPT-3.5. 

\begin{figure}[t]
  \centering
  \includegraphics[width=\linewidth]{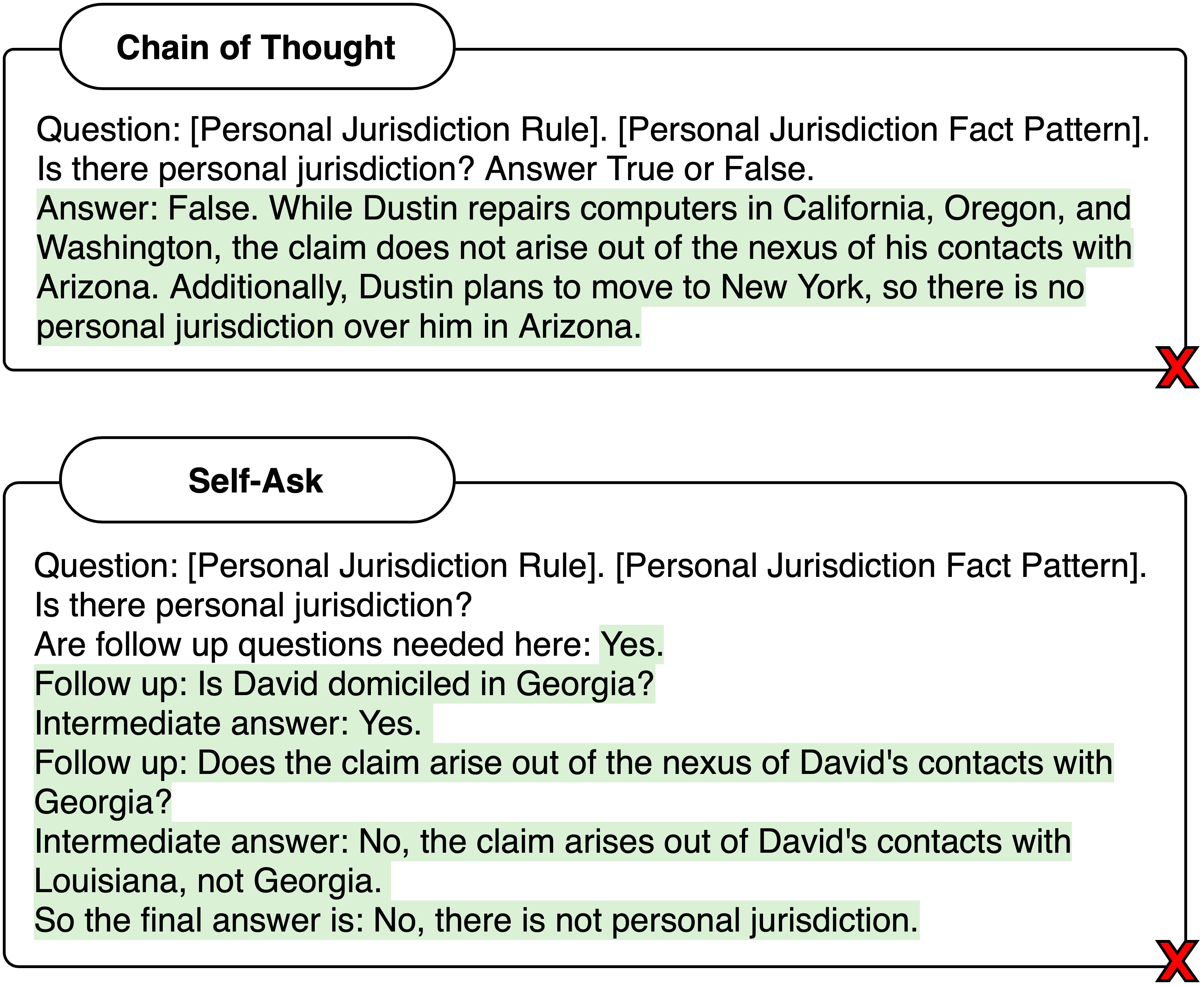}
  \caption{GPT-4 output demonstrating common errors in one-shot methods.}
  \label{figure:errors}
\end{figure}

The one-shot baselines demonstrate the challenges to in-context learning in this setting by consistently underperforming zero-shot methods. The model output depicted in Figures \ref{figure:output} and \ref{figure:errors} illustrate common errors we observe in rule-based reasoning using chain of thought and self-ask. First, chain of thought often generates fluent but illogical reasoning paths. The output in Figure \ref{figure:errors} shows the model incorrectly conclude that personal jurisdiction does not exist because Dustin plans to move to New York, even though future residency is not an element of the rule. Second, the question/answer pairs generated by the self-ask approach commonly decompose the rule incorrectly (Figure \ref{figure:output}) or incompletely (Figure \ref{figure:errors}). The output in the latter figure shows the model only asks intermediate questions for two out of three rule elements, omitting whether the defendant had sufficient contact with the forum state. Chain of logic prevents these omissions by first restating the rule elements (step 2 in the approach), similar to how System 2 Attention \cite{weston20232} first restates only the relevant portions of a context before attending to a task. In the ablation study, we show that removing this step significantly decreases performance. Last, Figure \ref{figure:errors} also demonstrates how not recognizing logical relationships can lead to an incorrect solution, even when sub-questions are answered correctly. The model correctly determines the defendant is domiciled in the forum state, which is sufficient to conclude personal jurisdiction exists, but still reaches the wrong final answer.

\subsection{Diversity Jurisdiction Series} \label{diversity_series}

The Diversity Jurisdiction series gives us a unique opportunity to examine how performance is affected by task complexity. Recall there are 6 tasks involving the diversity jurisdiction rule. The fact patterns in these tasks increase in complexity across two dimensions: 1) the number of parties (plaintiffs and defendants), and the number of claims per plaintiff-defendant pair. The simplest task, Diversity Jurisdiction 1 (DJ1), contains one plaintiff, one defendant and one claim. The most complex task, Diversity Jurisdiction 6 (DJ6), contains two plaintiffs, two defendants, and 4 total claims. 

Figure \ref{figure:diversity} shows results across all 6 Diversity Jurisdiction tasks using GPT-3.5. All methods, except zero-shot-LR, perform above 95\% accuracy on DJ1, yet even the best performing method, chain of logic, only scores a 61.7\% accuracy on DJ6. Chain of logic outperforms the baseline methods for 5 out of 6 tasks, with standard prompting outperforming our method by 2.3\% for DJ2. Most notably, chain of logic significantly outperforms other methods on tasks requiring arithmetic reasoning to add dollar amounts from multiple claims (DJ3, DJ5, DJ6). The performance gap between chain of logic and zero-shot, the best performing baseline, is 36\% for DJ5. All baselines perform near random chance for this task. For tasks involving arithmetic reasoning, the baselines also demonstrate an inverse relationship between prompt complexity and performance. That is to say, simpler prompts tend to perform better. Chain of logic is the clear exception to this trend being both the most complex prompt and the best performing model.

\subsection{Ablation Study}

Our experiments show chain of logic improves in-context learning for rule-based reasoning tasks. In this section, we assess the contribution of each reasoning step to the overall performance as shown in Table \ref{tab:ablations}. We isolate these effects by removing a single reasoning step at a time. We evaluate on the most complex rule, personal jurisdiction, using the model with the best comparable performance, GPT-3.5. Unsurprisingly, removing step 4 drastically reduces overall performance since there is no clear path between the logical expressions in steps 3 and 5.  Notably, the removal of step 1, switching from structured to unstructured task inputs, leads to a 25.9 absolute percentage point performance decrease. We believe this structured content is useful for the rule decomposition step that follows. Similarly, the ablations show that decomposing the rule into elements (step 2: -12.9\%) and generating a logical expression representing the relationships between elements (step 3: -14.8\%) are both critical to the performance gains we observe for this approach.

\begin{table}
  \centering
  \begin{tabular}{lr}
    \toprule
    Reasoning Step & PJ \\ \midrule
    \textbf{Step 1:} Structured Input & -25.9 \\
    \textbf{Step 2:} Rule Decomposition & -12.9 \\
    \textbf{Step 3:} Logical Expression & -14.8 \\
    \textbf{Step 4:} Question Answering & -49.3 \\
    \textbf{Step 5:} Element Recomposition & -5.5 \\
    \textbf{Step 6:} Resolve Expression & -3.7 \\
    \bottomrule
  \end{tabular}
  \caption{\textbf{Ablations.} Reduction in performance (absolute percentage) observed when removing the specified reasoning step from the chain of logic approach using GPT-3.5 on the Personal Jurisdiction task.}
  \label{tab:ablations}
\end{table}

\section{Related Work}

There has been extensive work on improving the problem solving capabilities of LMs by enabling a model to use more computation for more complex tasks. Graves introduces an adaptive algorithm for dynamically selecting the number of computation steps a recurrent neural network should take based on the task complexity \cite{graves2017adaptive}. Ling et al. first demonstrated the use of answer rationales, natural language explanations generated before the solution, to solve algebraic word problems \cite{ling2017program}. Several approaches followed this work in exploring the use of intermediate reasoning steps \cite{nye2021work, wei2023chainofthought, wang2023selfconsistency}. More similar to our work, other methods have explored decomposing compositional questions into simpler sub-questions using supervised models \cite{qi2019answering, min2019multihop, khot2021text} and prompting techniques \cite{mishra2022reframing, press2023measuring}. Talmor and Berant also recognized the importance of explicitly identifying relationships between sub-questions \cite{talmor2018web}. The authors train a supervised model to translate questions into a computation tree and answer sub-questions by querying a search engine. We extend these works with our decomposition-recomposition approach, exploring simple questions involving compositional rules where the solution depends not only on rule element answers, but also the logical relationships between elements.

Prior work in natural language processing has explored a broad range of legal tasks, including clause classification \cite{hendrycks2021cuad} and recommendation \cite{aggarwal-etal-2021-clauserec}, contract summarization \cite{manor-li-2019-plain}, legal judgment prediction \cite{chalkidis2019neural} and even answering bar exam questions \cite{bommarito2022gpt, zhong2019jecqa}. Rule-based reasoning can involve rules from many sources. Saeidi et al. explored conversational question answering where the solution required rule-based reasoning about a collection of regulations \cite{saeidi2018interpretation}. Servantez et al. captured rules in contract text through graph-based extraction and converted them into code \cite{10.1145/3594536.3595162}. Similar to our work, Holzenberger and Van Durme introduced an approach to reasoning about tax statutes by decomposing the reasoning process \cite{holzenberger-van-durme-2021-factoring}. This approach first extracts key arguments (entities, dates, dollar amounts) from the statute text and fact pattern using fine-tuned BERT models, before arriving at a final answer. We do not explore this method since it requires a fine-tuned model and does not easily translate to the tasks in this work.

\section{Limitations and Future Work}

While LegalBench tasks were crafted by subject matter experts, the rules have been simplified to ensure answers are objectively correct. This greatly improves evaluation, but also means the scope of complexity is narrower than many real world rules. Additionally, our current approach only addresses rules where the solution is based on whether the antecedent has been triggered. Real world rules can contain complex consequences which themselves require some form of reasoning. For example, calculating tax liability after determining the applicable tax bracket. In future work, we hope to build on our current approach in several directions. First, for simplicity chain of logic performs all reasoning step in a single forward pass. However, a multiple pass approach would allow us to incorporate other reasoning tasks like rule identification which is not addressed here. This moves us toward more real world scenarios where the applicable rule is not known a priori. Second, investigating whether rule-based reasoning could be improved further by dynamically sampling multiple reasoning paths. And third, incorporating retrieval augmented generation \cite{lewis2021retrievalaugmented} by allowing the model to access external sources of knowledge like term definitions. This is particularly useful in the legal domain where terms or concepts have a distinct meaning based on the jurisdiction or contract.

\section{Conclusion}
In this work, we explore causal language models as rule-based reasoners, and show these models can have difficulty learning from in-context demonstrations using common prompting approaches. We present a new prompting method, chain of logic, to elicit rule-based reasoning in language models through decomposition and recomposition. Our experiments show chain of logic consistently outperforms other prompting methods, including chain of thought and self-ask, across a variety of rule-based reasoning tasks using both open-source and commercial language models. We show how chain of logic creates a coherent and interpretable reasoning path by unraveling the many reasoning steps required by compositional rules. These findings may also be a useful signal for future instruction tuning of language models to imbue them with reasoning. By enhancing the rule-based reasoning capabilities of LMs through in-context learning, chain of logic also reduces the need for annotated legal data, which has historically been a bottleneck for the legal domain.

\section*{Acknowledgements}
This work was supported in part by Adobe Research. The authors thank Neel Guha for helpful discussions related to LegalBench.

\bibliography{anthology,references}

\begin{thebibliography}{37}
\expandafter\ifx\csname natexlab\endcsname\relax\def\natexlab#1{#1}\fi

\bibitem[{Aggarwal et~al.(2021)Aggarwal, Garimella, Srinivasan, N, and Jain}]{aggarwal-etal-2021-clauserec}
Vinay Aggarwal, Aparna Garimella, Balaji~Vasan Srinivasan, Anandhavelu N, and Rajiv Jain. 2021.
\newblock \href {https://doi.org/10.18653/v1/2021.emnlp-main.691} {{C}lause{R}ec: A clause recommendation framework for {AI}-aided contract authoring}.
\newblock In \emph{Proceedings of the 2021 Conference on Empirical Methods in Natural Language Processing}, pages 8770--8776, Online and Punta Cana, Dominican Republic. Association for Computational Linguistics.

\bibitem[{au2 and Katz(2022)}]{bommarito2022gpt}
Michael Bommarito~II au2 and Daniel~Martin Katz. 2022.
\newblock \href {http://arxiv.org/abs/2212.14402} {Gpt takes the bar exam}.

\bibitem[{Blair-Stanek et~al.(2023{\natexlab{a}})Blair-Stanek, Holzenberger, and Durme}]{blairstanek2023blt}
Andrew Blair-Stanek, Nils Holzenberger, and Benjamin~Van Durme. 2023{\natexlab{a}}.
\newblock \href {http://arxiv.org/abs/2311.09693} {Blt: Can large language models handle basic legal text?}

\bibitem[{Blair-Stanek et~al.(2023{\natexlab{b}})Blair-Stanek, Holzenberger, and Van~Durme}]{10.1145/3594536.3595163}
Andrew Blair-Stanek, Nils Holzenberger, and Benjamin Van~Durme. 2023{\natexlab{b}}.
\newblock \href {https://doi.org/10.1145/3594536.3595163} {Can gpt-3 perform statutory reasoning?}
\newblock In \emph{Proceedings of the Nineteenth International Conference on Artificial Intelligence and Law}, ICAIL '23, page 22–31, New York, NY, USA. Association for Computing Machinery.

\bibitem[{Brown et~al.(2020)Brown, Mann, Ryder, Subbiah, Kaplan, Dhariwal, Neelakantan, Shyam, Sastry, Askell, Agarwal, Herbert-Voss, Krueger, Henighan, Child, Ramesh, Ziegler, Wu, Winter, Hesse, Chen, Sigler, Litwin, Gray, Chess, Clark, Berner, McCandlish, Radford, Sutskever, and Amodei}]{brown2020language}
Tom~B. Brown, Benjamin Mann, Nick Ryder, Melanie Subbiah, Jared Kaplan, Prafulla Dhariwal, Arvind Neelakantan, Pranav Shyam, Girish Sastry, Amanda Askell, Sandhini Agarwal, Ariel Herbert-Voss, Gretchen Krueger, Tom Henighan, Rewon Child, Aditya Ramesh, Daniel~M. Ziegler, Jeffrey Wu, Clemens Winter, Christopher Hesse, Mark Chen, Eric Sigler, Mateusz Litwin, Scott Gray, Benjamin Chess, Jack Clark, Christopher Berner, Sam McCandlish, Alec Radford, Ilya Sutskever, and Dario Amodei. 2020.
\newblock \href {http://arxiv.org/abs/2005.14165} {Language models are few-shot learners}.

\bibitem[{Chalkidis et~al.(2019)Chalkidis, Androutsopoulos, and Aletras}]{chalkidis2019neural}
Ilias Chalkidis, Ion Androutsopoulos, and Nikolaos Aletras. 2019.
\newblock \href {http://arxiv.org/abs/1906.02059} {Neural legal judgment prediction in english}.

\bibitem[{Dahl et~al.(2024)Dahl, Magesh, Suzgun, and Ho}]{dahl2024large}
Matthew Dahl, Varun Magesh, Mirac Suzgun, and Daniel~E. Ho. 2024.
\newblock \href {http://arxiv.org/abs/2401.01301} {Large legal fictions: Profiling legal hallucinations in large language models}.

\bibitem[{Graves(2017)}]{graves2017adaptive}
Alex Graves. 2017.
\newblock \href {http://arxiv.org/abs/1603.08983} {Adaptive computation time for recurrent neural networks}.

\bibitem[{Guha et~al.(2023)Guha, Nyarko, Ho, Ré, Chilton, Narayana, Chohlas-Wood, Peters, Waldon, Rockmore, Zambrano, Talisman, Hoque, Surani, Fagan, Sarfaty, Dickinson, Porat, Hegland, Wu, Nudell, Niklaus, Nay, Choi, Tobia, Hagan, Ma, Livermore, Rasumov-Rahe, Holzenberger, Kolt, Henderson, Rehaag, Goel, Gao, Williams, Gandhi, Zur, Iyer, and Li}]{guha2023legalbench}
Neel Guha, Julian Nyarko, Daniel~E. Ho, Christopher Ré, Adam Chilton, Aditya Narayana, Alex Chohlas-Wood, Austin Peters, Brandon Waldon, Daniel~N. Rockmore, Diego Zambrano, Dmitry Talisman, Enam Hoque, Faiz Surani, Frank Fagan, Galit Sarfaty, Gregory~M. Dickinson, Haggai Porat, Jason Hegland, Jessica Wu, Joe Nudell, Joel Niklaus, John Nay, Jonathan~H. Choi, Kevin Tobia, Margaret Hagan, Megan Ma, Michael Livermore, Nikon Rasumov-Rahe, Nils Holzenberger, Noam Kolt, Peter Henderson, Sean Rehaag, Sharad Goel, Shang Gao, Spencer Williams, Sunny Gandhi, Tom Zur, Varun Iyer, and Zehua Li. 2023.
\newblock \href {http://arxiv.org/abs/2308.11462} {Legalbench: A collaboratively built benchmark for measuring legal reasoning in large language models}.

\bibitem[{Hendrycks et~al.(2021)Hendrycks, Burns, Chen, and Ball}]{hendrycks2021cuad}
Dan Hendrycks, Collin Burns, Anya Chen, and Spencer Ball. 2021.
\newblock \href {http://arxiv.org/abs/2103.06268} {Cuad: An expert-annotated nlp dataset for legal contract review}.

\bibitem[{Holzenberger and Van~Durme(2021)}]{holzenberger-van-durme-2021-factoring}
Nils Holzenberger and Benjamin Van~Durme. 2021.
\newblock \href {https://doi.org/10.18653/v1/2021.acl-long.213} {Factoring statutory reasoning as language understanding challenges}.
\newblock In \emph{Proceedings of the 59th Annual Meeting of the Association for Computational Linguistics and the 11th International Joint Conference on Natural Language Processing (Volume 1: Long Papers)}, pages 2742--2758, Online. Association for Computational Linguistics.

\bibitem[{Jiang et~al.(2023)Jiang, Sablayrolles, Mensch, Bamford, Chaplot, de~las Casas, Bressand, Lengyel, Lample, Saulnier, Lavaud, Lachaux, Stock, Scao, Lavril, Wang, Lacroix, and Sayed}]{jiang2023mistral}
Albert~Q. Jiang, Alexandre Sablayrolles, Arthur Mensch, Chris Bamford, Devendra~Singh Chaplot, Diego de~las Casas, Florian Bressand, Gianna Lengyel, Guillaume Lample, Lucile Saulnier, Lélio~Renard Lavaud, Marie-Anne Lachaux, Pierre Stock, Teven~Le Scao, Thibaut Lavril, Thomas Wang, Timothée Lacroix, and William~El Sayed. 2023.
\newblock \href {http://arxiv.org/abs/2310.06825} {Mistral 7b}.

\bibitem[{Jiang and Yang(2023)}]{jiang2023legal}
Cong Jiang and Xiaolei Yang. 2023.
\newblock \href {http://arxiv.org/abs/2307.08321} {Legal syllogism prompting: Teaching large language models for legal judgment prediction}.

\bibitem[{Khot et~al.(2021)Khot, Khashabi, Richardson, Clark, and Sabharwal}]{khot2021text}
Tushar Khot, Daniel Khashabi, Kyle Richardson, Peter Clark, and Ashish Sabharwal. 2021.
\newblock \href {http://arxiv.org/abs/2009.00751} {Text modular networks: Learning to decompose tasks in the language of existing models}.

\bibitem[{Kojima et~al.(2023)Kojima, Gu, Reid, Matsuo, and Iwasawa}]{kojima2023large}
Takeshi Kojima, Shixiang~Shane Gu, Machel Reid, Yutaka Matsuo, and Yusuke Iwasawa. 2023.
\newblock \href {http://arxiv.org/abs/2205.11916} {Large language models are zero-shot reasoners}.

\bibitem[{Lewis et~al.(2021)Lewis, Perez, Piktus, Petroni, Karpukhin, Goyal, Küttler, Lewis, tau Yih, Rocktäschel, Riedel, and Kiela}]{lewis2021retrievalaugmented}
Patrick Lewis, Ethan Perez, Aleksandra Piktus, Fabio Petroni, Vladimir Karpukhin, Naman Goyal, Heinrich Küttler, Mike Lewis, Wen tau Yih, Tim Rocktäschel, Sebastian Riedel, and Douwe Kiela. 2021.
\newblock \href {http://arxiv.org/abs/2005.11401} {Retrieval-augmented generation for knowledge-intensive nlp tasks}.

\bibitem[{Ling et~al.(2017)Ling, Yogatama, Dyer, and Blunsom}]{ling2017program}
Wang Ling, Dani Yogatama, Chris Dyer, and Phil Blunsom. 2017.
\newblock \href {http://arxiv.org/abs/1705.04146} {Program induction by rationale generation : Learning to solve and explain algebraic word problems}.

\bibitem[{Manor and Li(2019)}]{manor-li-2019-plain}
Laura Manor and Junyi~Jessy Li. 2019.
\newblock \href {https://doi.org/10.18653/v1/W19-2201} {Plain {E}nglish summarization of contracts}.
\newblock In \emph{Proceedings of the Natural Legal Language Processing Workshop 2019}, pages 1--11, Minneapolis, Minnesota. Association for Computational Linguistics.

\bibitem[{Min et~al.(2019)Min, Zhong, Zettlemoyer, and Hajishirzi}]{min2019multihop}
Sewon Min, Victor Zhong, Luke Zettlemoyer, and Hannaneh Hajishirzi. 2019.
\newblock \href {http://arxiv.org/abs/1906.02916} {Multi-hop reading comprehension through question decomposition and rescoring}.

\bibitem[{Mishra et~al.(2022)Mishra, Khashabi, Baral, Choi, and Hajishirzi}]{mishra2022reframing}
Swaroop Mishra, Daniel Khashabi, Chitta Baral, Yejin Choi, and Hannaneh Hajishirzi. 2022.
\newblock \href {http://arxiv.org/abs/2109.07830} {Reframing instructional prompts to gptk's language}.

\bibitem[{Mukherjee et~al.(2023)Mukherjee, Mitra, Jawahar, Agarwal, Palangi, and Awadallah}]{mukherjee2023orca}
Subhabrata Mukherjee, Arindam Mitra, Ganesh Jawahar, Sahaj Agarwal, Hamid Palangi, and Ahmed Awadallah. 2023.
\newblock Orca: Progressive learning from complex explanation traces of gpt-4.
\newblock \emph{arXiv preprint arXiv:2306.02707}.

\bibitem[{Nye et~al.(2021)Nye, Andreassen, Gur-Ari, Michalewski, Austin, Bieber, Dohan, Lewkowycz, Bosma, Luan, Sutton, and Odena}]{nye2021work}
Maxwell Nye, Anders~Johan Andreassen, Guy Gur-Ari, Henryk Michalewski, Jacob Austin, David Bieber, David Dohan, Aitor Lewkowycz, Maarten Bosma, David Luan, Charles Sutton, and Augustus Odena. 2021.
\newblock \href {http://arxiv.org/abs/2112.00114} {Show your work: Scratchpads for intermediate computation with language models}.

\bibitem[{OpenAI(2023)}]{openai2023gpt}
OpenAI. 2023.
\newblock Gpt-4 technical report.
\newblock \emph{View in Article}, 2:13.

\bibitem[{Press et~al.(2023)Press, Zhang, Min, Schmidt, Smith, and Lewis}]{press2023measuring}
Ofir Press, Muru Zhang, Sewon Min, Ludwig Schmidt, Noah~A. Smith, and Mike Lewis. 2023.
\newblock \href {http://arxiv.org/abs/2210.03350} {Measuring and narrowing the compositionality gap in language models}.

\bibitem[{Qi et~al.(2019)Qi, Lin, Mehr, Wang, and Manning}]{qi2019answering}
Peng Qi, Xiaowen Lin, Leo Mehr, Zijian Wang, and Christopher~D. Manning. 2019.
\newblock \href {http://arxiv.org/abs/1910.07000} {Answering complex open-domain questions through iterative query generation}.

\bibitem[{Roth(1995)}]{roth1995law}
Martha~T Roth. 1995.
\newblock \emph{Law collections from Mesopotamia and Asia minor}, volume~6.
\newblock Scholars Press.

\bibitem[{Saeidi et~al.(2018)Saeidi, Bartolo, Lewis, Singh, Rocktäschel, Sheldon, Bouchard, and Riedel}]{saeidi2018interpretation}
Marzieh Saeidi, Max Bartolo, Patrick Lewis, Sameer Singh, Tim Rocktäschel, Mike Sheldon, Guillaume Bouchard, and Sebastian Riedel. 2018.
\newblock \href {http://arxiv.org/abs/1809.01494} {Interpretation of natural language rules in conversational machine reading}.

\bibitem[{Servantez et~al.(2023)Servantez, Lipka, Siu, Aggarwal, Krishnamurthy, Garimella, Hammond, and Jain}]{10.1145/3594536.3595162}
Sergio Servantez, Nedim Lipka, Alexa Siu, Milan Aggarwal, Balaji Krishnamurthy, Aparna Garimella, Kristian Hammond, and Rajiv Jain. 2023.
\newblock \href {https://doi.org/10.1145/3594536.3595162} {Computable contracts by extracting obligation logic graphs}.
\newblock In \emph{Proceedings of the Nineteenth International Conference on Artificial Intelligence and Law}, ICAIL '23, page 267–276, New York, NY, USA. Association for Computing Machinery.

\bibitem[{Talmor and Berant(2018)}]{talmor2018web}
Alon Talmor and Jonathan Berant. 2018.
\newblock \href {http://arxiv.org/abs/1803.06643} {The web as a knowledge-base for answering complex questions}.

\bibitem[{Touvron et~al.(2023)Touvron, Martin, Stone, Albert, Almahairi, Babaei, Bashlykov, Batra, Bhargava, Bhosale, Bikel, Blecher, Ferrer, Chen, Cucurull, Esiobu, Fernandes, Fu, Fu, Fuller, Gao, Goswami, Goyal, Hartshorn, Hosseini, Hou, Inan, Kardas, Kerkez, Khabsa, Kloumann, Korenev, Koura, Lachaux, Lavril, Lee, Liskovich, Lu, Mao, Martinet, Mihaylov, Mishra, Molybog, Nie, Poulton, Reizenstein, Rungta, Saladi, Schelten, Silva, Smith, Subramanian, Tan, Tang, Taylor, Williams, Kuan, Xu, Yan, Zarov, Zhang, Fan, Kambadur, Narang, Rodriguez, Stojnic, Edunov, and Scialom}]{touvron2023llama}
Hugo Touvron, Louis Martin, Kevin Stone, Peter Albert, Amjad Almahairi, Yasmine Babaei, Nikolay Bashlykov, Soumya Batra, Prajjwal Bhargava, Shruti Bhosale, Dan Bikel, Lukas Blecher, Cristian~Canton Ferrer, Moya Chen, Guillem Cucurull, David Esiobu, Jude Fernandes, Jeremy Fu, Wenyin Fu, Brian Fuller, Cynthia Gao, Vedanuj Goswami, Naman Goyal, Anthony Hartshorn, Saghar Hosseini, Rui Hou, Hakan Inan, Marcin Kardas, Viktor Kerkez, Madian Khabsa, Isabel Kloumann, Artem Korenev, Punit~Singh Koura, Marie-Anne Lachaux, Thibaut Lavril, Jenya Lee, Diana Liskovich, Yinghai Lu, Yuning Mao, Xavier Martinet, Todor Mihaylov, Pushkar Mishra, Igor Molybog, Yixin Nie, Andrew Poulton, Jeremy Reizenstein, Rashi Rungta, Kalyan Saladi, Alan Schelten, Ruan Silva, Eric~Michael Smith, Ranjan Subramanian, Xiaoqing~Ellen Tan, Binh Tang, Ross Taylor, Adina Williams, Jian~Xiang Kuan, Puxin Xu, Zheng Yan, Iliyan Zarov, Yuchen Zhang, Angela Fan, Melanie Kambadur, Sharan Narang, Aurelien Rodriguez, Robert Stojnic, Sergey Edunov, and Thomas
  Scialom. 2023.
\newblock \href {http://arxiv.org/abs/2307.09288} {Llama 2: Open foundation and fine-tuned chat models}.

\bibitem[{Wang et~al.(2023)Wang, Wei, Schuurmans, Le, Chi, Narang, Chowdhery, and Zhou}]{wang2023selfconsistency}
Xuezhi Wang, Jason Wei, Dale Schuurmans, Quoc Le, Ed~Chi, Sharan Narang, Aakanksha Chowdhery, and Denny Zhou. 2023.
\newblock \href {http://arxiv.org/abs/2203.11171} {Self-consistency improves chain of thought reasoning in language models}.

\bibitem[{Wei et~al.(2023)Wei, Wang, Schuurmans, Bosma, Ichter, Xia, Chi, Le, and Zhou}]{wei2023chainofthought}
Jason Wei, Xuezhi Wang, Dale Schuurmans, Maarten Bosma, Brian Ichter, Fei Xia, Ed~Chi, Quoc Le, and Denny Zhou. 2023.
\newblock \href {http://arxiv.org/abs/2201.11903} {Chain-of-thought prompting elicits reasoning in large language models}.

\bibitem[{Weston and Sukhbaatar(2023)}]{weston20232}
Jason Weston and Sainbayar Sukhbaatar. 2023.
\newblock \href {http://arxiv.org/abs/2311.11829} {System 2 attention (is something you might need too)}.

\bibitem[{Wolfson et~al.(2020)Wolfson, Geva, Gupta, Gardner, Goldberg, Deutch, and Berant}]{wolfson2020break}
Tomer Wolfson, Mor Geva, Ankit Gupta, Matt Gardner, Yoav Goldberg, Daniel Deutch, and Jonathan Berant. 2020.
\newblock \href {http://arxiv.org/abs/2001.11770} {Break it down: A question understanding benchmark}.

\bibitem[{Yu et~al.(2022)Yu, Quartey, and Schilder}]{yu2022legal}
Fangyi Yu, Lee Quartey, and Frank Schilder. 2022.
\newblock \href {http://arxiv.org/abs/2212.01326} {Legal prompting: Teaching a language model to think like a lawyer}.

\bibitem[{Zhong et~al.(2019)Zhong, Xiao, Tu, Zhang, Liu, and Sun}]{zhong2019jecqa}
Haoxi Zhong, Chaojun Xiao, Cunchao Tu, Tianyang Zhang, Zhiyuan Liu, and Maosong Sun. 2019.
\newblock \href {http://arxiv.org/abs/1911.12011} {Jec-qa: A legal-domain question answering dataset}.

\bibitem[{Zhou et~al.(2023)Zhou, Schärli, Hou, Wei, Scales, Wang, Schuurmans, Cui, Bousquet, Le, and Chi}]{zhou2023leasttomost}
Denny Zhou, Nathanael Schärli, Le~Hou, Jason Wei, Nathan Scales, Xuezhi Wang, Dale Schuurmans, Claire Cui, Olivier Bousquet, Quoc Le, and Ed~Chi. 2023.
\newblock \href {http://arxiv.org/abs/2205.10625} {Least-to-most prompting enables complex reasoning in large language models}.

\end{thebibliography}
\bibliographystyle{acl_natbib}

\appendix

\section{Appendix}
\label{sec:appendix}

\subsection{LegalBench Task Examples}
\textbf{Personal Jurisdiction} (Figure \ref{figure:output})
\begin{itemize}
\item \underline{Rule}: There is personal jurisdiction over a defendant in the state where the defendant is domiciled, or when (1) the defendant has sufficient contacts with the state, such that they have availed itself of the privileges of the state and (2) the claim arises out of the nexus of the defendant's contacts with the state.
\item \underline{Fact Pattern}: Dustin is a repairman who lives in Arizona and repairs computers in California, Oregon, and Washington. While travelling to repair a computer in Washington, Dustin is involved in a car crash in Oregon with Laura, a citizen of Texas. After the accident, Dustin returns to Arizona. Laura sues him in Oregon.  
\item \underline{Issue}: Is there personal jurisdiction? 
\end{itemize}

\textbf{Diversity Jurisdiction 1}
\begin{itemize}
\item \underline{Rule}: Diversity jurisdiction exists when there is (1) complete diversity between plaintiffs and defendants, and (2) the amount-in-controversy (AiC) is greater than \$75k.
\item \underline{Fact Pattern}: James is from Arizona. Lucas is from Arizona. James sues Lucas for negligence for \$64,000.
\item \underline{Issue}: Is there diversity jurisdiction?
\end{itemize}

\textbf{Diversity Jurisdiction 2}
\begin{itemize}
\item \underline{Rule}: Diversity jurisdiction exists when there is (1) complete diversity between plaintiffs and defendants, and (2) the amount-in-controversy (AiC) is greater than \$75k.
\item \underline{Fact Pattern}: Sophia is from Arkansas. Benjamin is from Hawaii. Noah is from Arkansas. Sophia sues Benjamin and Noah each for defamation for \$24,000.
\item \underline{Issue}: Is there diversity jurisdiction?
\end{itemize}

\textbf{Diversity Jurisdiction 3}
\begin{itemize}
\item \underline{Rule}: Diversity jurisdiction exists when there is (1) complete diversity between plaintiffs and defendants, and (2) the amount-in-controversy (AiC) is greater than \$75k.
\item \underline{Fact Pattern}: William is from Montana. Theodore is from Connecticut. William sues Theodore for medical malpractice for \$9,000 and negligence for \$35,000.
\item \underline{Issue}: Is there diversity jurisdiction?
\end{itemize}

\textbf{Diversity Jurisdiction 4}
\begin{itemize}
\item \underline{Rule}: Diversity jurisdiction exists when there is (1) complete diversity between plaintiffs and defendants, and (2) the amount-in-controversy (AiC) is greater than \$75k.
\item \underline{Fact Pattern}: Emma is from New Hampshire. Mia is from Wisconsin. Evelyn is from California. Emma and Mia both sue Evelyn for copyright infringement for \$5,400,000.
\item \underline{Issue}: Is there diversity jurisdiction?
\end{itemize}

\textbf{Diversity Jurisdiction 5}
\begin{itemize}
\item \underline{Rule}: Diversity jurisdiction exists when there is (1) complete diversity between plaintiffs and defendants, and (2) the amount-in-controversy (AiC) is greater than \$75k.
\item \underline{Fact Pattern}: Elijah is from Hawaii. Ava is from Oklahoma. Amelia is from Minnesota. Elijah and Ava both sue Amelia for defamation for \$3,000 and copyright infringement for \$80,000.
\item \underline{Issue}: Is there diversity jurisdiction?
\end{itemize}

\textbf{Diversity Jurisdiction 6}
\begin{itemize}
\item \underline{Rule}: Diversity jurisdiction exists when there is (1) complete diversity between plaintiffs and defendants, and (2) the amount-in-controversy (AiC) is greater than \$75k.
\item \underline{Fact Pattern}: Theodore is from North Dakota. Amelia is from Georgia. Benjamin is from Delaware. Mia is from Illinois. Theodore and Amelia both sue Benjamin for trademark infringement for \$42,000 and copyright infringement for \$71,000. Theodore and Amelia both sue Mia for securities fraud for \$45,000 and medical malpractice for \$57,000.
\item \underline{Issue}: Is there diversity jurisdiction?
\end{itemize}

\textbf{J.Crew Blocker}
\begin{itemize}
\item \underline{Rule}: The JCrew Blocker is a provision that typically includes (1) a prohibition on the borrower from transferring IP to an unrestricted subsidiary, and (2) a requirement that the borrower obtains the consent of its agent/lenders before transferring IP to any subsidiary.
\item \underline{Fact Pattern}: Notwithstanding anything to foregoing, no Intellectual Property that is material to the Borrower and its Restricted Subsidiaries, taken as a whole (as reasonably determined by the Borrower), shall be owned by or licensed, contributed or otherwise transferred to any Unrestricted Subsidiary.
\item \underline{Issue}: Do the following provisions contain JCrew Blockers?
\end{itemize}

\subsection{Model Performance by Rule}
Model performance (accuracy) by rule for all reasoning tasks. Diversity Jurisdiction results are aggregated across all six datasets.

\begin{table*}[t]
  \centering
  \label{tab:results_gpt35}
  \begin{tabular}{ lrrr }
    \toprule
     & Personal Jurisdiction & Diversity Jurisdiction & J.Crew Blocker \\ \midrule
    Zero-Shot & 68.0 & 73.9 & 87.0 \\ 
     Zero-Shot-LR & 60.0 & 69.2 & 42.6 \\
    Zero-Shot-LS & 70.0  & 69.5 & 87.0  \\
    \midrule
    Standard Prompting & 72.0 & 73.5 & 72.2 \\
    Chain of Thought & 74.0 & 69.7 & 66.7 \\
    Self-Ask & 60.0 & 70.7 & 74.0 \\
    Chain of Logic (ours) & \textbf{78.0} & \textbf{88.6} & \textbf{94.4} \\
    \bottomrule
  \end{tabular}
  \caption{GPT-3.5 Performance}
\end{table*}

\begin{table*}[t]
  \centering
  \label{tab:results_gpt4}
  \begin{tabular}{ lrrr }
    \toprule
     & Personal Jurisdiction & Diversity Jurisdiction & J.Crew Blocker \\ \midrule
    Zero-Shot & 84.0 & 87.1 & \textbf{100} \\ 
    Zero-Shot-LR & 86.0 & 87.8 & 98.1\\
    Zero-Shot-LS & 78.0 & 94.2 & 98.1  \\
    \midrule
    Standard Prompting & 76.0 & 93.6 & 96.3 \\
    Chain of Thought & 82.0 & 89.2 & 96.3 \\
    Self-Ask & 84.0 & 78.5 & 96.3 \\
    Chain of Logic (ours) & \textbf{90.0} & \textbf{94.3} & 92.6 \\
    \bottomrule
  \end{tabular}
  \caption{GPT-4 Performance}
\end{table*}

\begin{table*}[t]
  \centering
  \label{tab:results_llama}
  \begin{tabular}{ lrrr }
    \toprule
     & Personal Jurisdiction & Diversity Jurisdiction & J.Crew Blocker \\ \midrule
    Zero-Shot & 66.0 & 65.9 & 90.7 \\ 
    Zero-Shot-LR & 52.0 & 66.0 & 77.8 \\
    Zero-Shot-LS & 42.0 & 50.0 & 83.3  \\
    \midrule
    Standard Prompting & 56.0 & 57.7 & 81.5 \\
    Chain of Thought & 62.0 & 60.7 & \textbf{92.6} \\
    Self-Ask & 70.0 & 52.8 & 79.6 \\
    Chain of Logic (ours) & \textbf{72.0} & \textbf{66.6} & 85.2 \\
    \bottomrule
  \end{tabular}
  \caption{Llama-2-70b Performance}
\end{table*}

\begin{table*}[t]
  \centering
  \label{tab:results_mistral}
  \begin{tabular}{ lrrr }
    \toprule
     & Personal Jurisdiction & Diversity Jurisdiction & J.Crew Blocker \\ \midrule
    Zero-Shot & \textbf{56.0} & 65.4 & 61.1 \\ 
    Zero-Shot-LR & 50.0 & 66.1 & \textbf{72.2} \\
    Zero-Shot-LS & 48.0 & 62.4 & 46.3  \\
    \midrule
    Standard Prompting & 44.0 & 46.3 & 46.3 \\
    Chain of Thought & 36.0 & 66.2 & 37.0 \\
    Self-Ask & 52.0 & 70.8 & 16.7 \\
    Chain of Logic (ours) & 54.0 & \textbf{72.4} & 63.0 \\
    \bottomrule
  \end{tabular}
  \caption{MistralOrca-7b Performance}
\end{table*}

\end{document}